# Hybrid SARIMA–LSTM Model for Local Weather Forecasting: A Residual-Learning Approach for Data-Driven Meteorological Prediction

Shreyas Rajeev
*Stevens Institute of Technology*
Hoboken, USA
srajeev@stevens.edu

Karthik Mudenahalli Ashoka
Stevens Institute of Technology
*Hoboken, USA*
kmudenah@stevens.edu

Amit Mallappa Tiparaddi
Stevens Institute of Technology
Hoboken, USA
aitiparad@stevens.edu

## I. INTRODUCTION

For decades, accurately predicting the weather over the long term has been a major scientific problem because atmospheric systems are naturally chaotic. Temperature shows how cyclical climate forces and short-term changes that are not very regular work together. Seasonal changes happen in a regular way because of the way the Earth moves, and its tilt, the way the ocean and atmosphere interact, and long-term weather patterns. These yearly temperature patterns usually show up as smooth, repeating cycles. But short-term changes, like warm fronts, cold air intrusions, pressure driven anomalies, and sudden changes in the humidity or wind conditions, cause nonlinear fluctuations that make it much harder to make accurate predictions.

SARIMA (Seasonal autoregressive integrated moving average) and other traditional statistical methods have been used a lot to model historical weather data because they make so much sense and show linear seasonal patterns. But these methods only work if the relationships in the data stay the same over time, and they cannot show nonlinear behavior very well. Because of this, models like SARIMA tend to underpredict sudden temperature rises or overpredict declines. Therefore, this leads to systematic residual errors that get worse when you try to predict beyond a short time frame.

Deep learning techniques, particularly recurrent neural networks like LSTMs (Long Short-Term Memory), have exhibited exceptional efficacy in addressing intricate time-series challenges by acquiring nonlinear temporal dependencies from data. LSTMs have memory gates that let them remember things over long sequences and find small patterns in multivariate data. But when LSTMs are asked to predict far into the future without any ground truth feedback, they become unstable because small errors in predictions spread and grow in recursive forecasting loops. This problem is especially bad when it comes to predicting the weather, where forecasts that last for weeks or months need to be based on a lot of predictions that are made repeatedly.

These limitations indicate that neither classical statistical models nor independent neural network models are adequate for reliable, precise long-term forecasting. This leads to a hybrid modeling approach that combines the best parts of both methods. We suggest a Hybrid SARIMA–LSTM architecture for this project that breaks temperature into a climate component that can be predicted and a weather component that is not linear. This is done with a residual-learning strategy: SARIMA models the long-term seasonal trend, and the LSTM model learns the random changes that happen in SARIMA's prediction errors.

We use Fourier seasonal encoding to clearly show the yearly cycle, and we add a new stabilized recursive forecasting mechanism that keeps predictions within a 293-day future horizon without any direct observation. The goal of this project is to use data from 2020 to 2023 to accurately predict the daily average temperature in New York City and to see if the hybrid model can do much better than SARIMA and LSTM on their own. This report fully describes the dataset, the related literature, the model's design and implementation, the experimental analysis, and the conclusions reached from this research.

## II. RELATED WORK

Recently, there has been a lot of research on time series forecasting for weather and environmental data, especially in the hybrid modeling frameworks that use both statistical and deep learning methods. The SARIMA and LSTM neural networks are the two examples of the models widely used because they can effectively capture the temporal dependencies, seasonal patterns, and nonlinearities often found in the meteorological datasets. Therefore, these models have been utilized in various fields, such as indoor climate control, regional weather forecasting, rainfall nowcasting, and renewable energy assessment. The literature consistently shows that both statistical and neural methods are useful and combining them often works better when datasets have both long-term seasonal patterns and short-term irregular changes.

Cao, Li, and Sun [1] created a SARIMAX–LSTM model that compares different models to predict the temperatures of the inner walls of homes in Xi'an, China. Their research showed that SARIMAX made very much of the accurate predictions when the system showed strong seasonal regularity. The MAPE for SARIMAX was 0.24%, while the MAPE for LSTM was 1.45%. Their results showed that statistical models work better than deep learning models when the time series doesn't change much in a nonlinear way and when outside variables are strongly related to the target. This finding underscores the enduring significance of SARIMA like models in the hybrid systems, especially in contexts characterized by stable climatic conditions. Kumari and Muthulakshmi [2] suggested a SARIMA based forecasting framework that uses real weather data like temperature, humidity, wind speed, and rainfall.

Their method put a lot of emphasis on careful preprocessing, which included differencing, normalization, and model-order selection using AIC and BIC, to make sure the results were more accurate. Their findings, with RMSE and MAE values of 0.87°C and 1.24°C, respectively, support the idea that SARIMA is still useful for short-term temperature prediction when the data is of high quality and does not change. Their conclusions are also in line with other evidence that shows that statistical models are still useful for small to medium datasets that need to be easy to understand.

He et al. [3] went beyond just using statistical models and created a hybrid deep-learning rainfall nowcasting model that combines LSTM with Decision Tree Regression. This model uses NSS- derived atmospheric variables like PWV and CAPE. They have improved their Deep Learning Rainfall Nowcast model beyond the older methods with an overall RMSE value of 1.25 mm (about 0.05 in) and an MAE of 0.37 mm (about 0.01 in). Besides that, it was also quite explicit in indicating which of the predictors were most significant. This research demonstrates that nonlinear deep learning models perform exceptionally well when meteorological processes are affected by dynamic atmospheric indicators instead of rigid seasonal patterns.

Hossain, Shams, and Ullah [4] conducted a comparative study of ARIMA, SARIMA, and LSTM in forecasting renewable energy, using ten years of solar and wind data. According to their results, SARIMA was able to capture only the basic temporal structure, but the LSTM model showed high predictive power, with average $R^2$ values greater than 0.98 during cross-validation. Their research emphasizes that LSTM is incredibly powerful when it comes to modeling non-linear relationships and for keeping long-term continuity, which is a characteristic of a large amount of data.

Krascsenits and Kiss [5] went even deeper into the deep learning advantages by comparing ARIMA and LSTM for temperature prediction in Budapest and Los Angeles. Their experiments revealed that LSTM models outperform ARIMA models in predicting short-term temperatures in areas where the weather is highly variable. This evidence the significant potential of nonlinear learning in areas with many different climates. Machado, Ataíde, and Borges [6] expanded on this premise by evaluating LSTM and BiLSTM architectures with meteorological data from multiple stations in Brazil. They concluded that BiLSTM achieves superior results as it can capture the temporal dependencies both in the forward and backward directions.

Chen et al. [7], on the other hand, looked at temperature data from Nanjing, China, over the course of more than six decades and found that SARIMA is still very reliable for long-term predictions when the climate series is periodic and stationary. Their findings underscore that, notwithstanding the emergence of deep learning, traditional models continue to possess significant forecasting utility in stable meteorological conditions. These studies collectively demonstrate a distinct methodological progression from exclusively statistical methods to deep learning and hybrid modeling. On all the fronts, studies have substantiated that SARIMA delivers very reliable and easy to comprehend forecasts for well-organized seasonal patterns, while LSTM models have a higher degree of flexibility in identifying non-linear behaviors and abrupt changes in the weather.

The idea of combining these views has made many researchers take the hybrid SARIMA-LSTM architectures that combine the advantages of statistical interpretability and neural adaptability as their way. Such hybrid models have been found to have higher accuracy, better stability over time, and greater robustness in a wide range of environmental prediction tasks. Hence, they constitute a robust theoretical backing for the hybrid residual-learning approach employed in this work.

### III. OUR SOLUTION

*A. Description of the Dataset:*

The data set used in this research is the result of a four-year-long (2020-2023) daily meteorological observation study in New York City, which was taken from the Open-Meteo Historical Weather Archive [8]. The length of this period is enough to show not only the repeated seasonal temperature patterns but also the short-term changes of the atmosphere which appear to be very random. The dataset comprises daily averages of temperature, dew point, atmospheric pressure, wind speed, and visibility, which, from a thermodynamic point of view, describe the local climate. We went for these measurements as they are very good surface air temperature indicators and at the same time give us some indirect information about weather patterns such as humidity cycles, pressure-driven storm systems and atmospheric clarity, which all impact temperature changes.

The Open-Meteo dataset is a good-quality data source with only a few preprocessing steps required to convert the raw data into a machine learning-friendly format. One of the problems with data related to weather is the issue of missing or partially incomplete readings. These may arise if sensors are broken, maintenance is not done, or data is delayed in transit. To bridge the gap in the time series, missing values were supplemented by forward- and backward-filling methods that maintain the time structure of the data without introducing fake noise. There was a slow drift effect of atmospheric pressure over the long term which caused the series to be non-stationary and thus unfit for training the LSTM network. To resolve this issue, the pressure points were differentiated which brought the series to a stable state and ensured that changes came from short-term weather patterns and not the accumulation of measurements.

MinMax normalization transformed all the variables in such a way that they were within the range of [0,1]. This made the model treat each feature equally regardless of its physical unit. Besides this, two Fourier terms - a sine component and a cosine component - were also derived because of the day of the year to represent the Earth's orbit around the sun in a cycle. These terms attract attention to the annual seasonal cycle very explicitly so that both SARIMA and LSTM can make use of seasonality without the need for large lag structures which would complicate the calculations. After finishing the preprocessing, the dataset was divided into training and testing subsets using a time criterion. About 80% of the data was used for building the model while the remaining 293 days were reserved for out-of-sample forecasting. This strict separation of time ensures that the model's performance is evaluated in a genuinely predictive setting, i.e. without any data leaks or unintentional future information.

The most recent version of the dataset is clean, well-structured, and multivariate, fitting a time series of study and thus very appropriate for the research questions of this project. It acts as a springboard for the sequential implementation and assessment of statistical modeling, deep learning, and hybrid forecasting strategies.

*B. Machine Learning Algorithms:*

The forecasting problem this project is working on is very hard because it requires being able to predict the temperature for almost ten months. For such a long prediction window, you need a model that can show both the long-term seasonal changes and the short-term nonlinear changes that are typical of urban temperature patterns. To meet these needs, the study uses a hybrid framework that combines a SARIMA model with a LSTM neural network. These algorithms have different fundamental assumptions and mathematical operations, but their strengths are complementary, which makes them suitable to be integrated in a single system.

SARIMA is a classic statistically based forecasting approach for modeling seasonal time series with autoregressive behavior. It indicates the value of a variable by linearly combining past observations, differenced terms, and seasonal lag components. This model accounts for the stable seasonal variations caused by the Earth's annual temperature cycle. Several studies have demonstrated that SARIMA can predict temperature trends accurately when the series is stationary and exhibits clear periodic patterns. SARIMA is a very good baseline model for capturing long-term patterns in New York City because the temperature follows a strong and predictable yearly rhythm. However, SARIMA's linear formulation limits its ability to model abrupt temperature changes resulting from non-linear atmospheric forces.

Conversely, LSTM networks were designed to locate nonlinear time dependencies in sequential data. The LSTM cells have memory gates that regulate how much information is retained or discarded over time, which is different from standard recurrent networks. Through this system, LSTM networks can understand complex relationships between the variables, which makes them very suitable for problems like weather forecasting, where a large number of weather variables simultaneously affect the target. LSTMs can identify tiny links in the changes of the dew point, pressure, and wind that are related to the sudden change of temperature. LSTMs hold this privilege, but they often commit errors quickly when they are required to provide long-term predictions without the use of real future data. This frailty renders it impossible to employ stand-alone LSTM models for multi-month forecasting unless they have some additional structural safeguards.

The hybrid model exploits the advantages of both approaches by dividing the prediction task into two: a linear climatic part represented by SARIMA and a nonlinear residual part captured by LSTM. SARIMA initially comes up with an estimate of the overall seasonal pattern of the temperature series. LSTM learns to generate the residuals of the forecast that are the target series, i.e., short-term changes in the prediction residuals that cannot be explained by the meteorological factors are reproduced by it with the help of multiple inputs. Thanks to this decomposition, LSTM is not required to learn the whole temperature function from scratch. Instead, it only learns those parts which are not modeled by SARIMA. The combination of SARIMA baseline and LSTM residuals yields a better prediction than each of these models separately.

This combined approach matches well with the current understanding of meteorological time series which argue that no single model can entirely explain them. The hybrid SARIMA-LSTM architecture is a thoughtful and experimentally verified solution to the problem of long-term temperature prediction.

*C. Implementation Details*

The hybrid forecasting model is technically the most complex part of this work. The system integrates data cleaning, feature engineering, statistical modeling, neural network training, and recursive multi-step forecasting into a unified, stable, and precise framework.

The very first step of the implementation is to set up the whole data processing pipeline. The Open-Meteo API is delivering raw data that is afterward transformed into a uniform numerical format. To ensure the smooth running of the process, missing entries are filled, and stabilization of atmospheric pressure is done. After that, the days of the year are represented with Fourier seasonal encoding to show that they are moving in a cycle. MinMax scaling brings all features to the same scale so that the LSTM's learning process is not affected by numerical magnitudes. After scaling, the data is converted into sequences suitable for supervised learning. Each training sample consists of a fourteen-day window of multivariate inputs to predict the SARIMA residual or temperature value for the next day.

The SARIMA component of the hybrid model is created by the SARIMAX class from the Python library statsmodels. The parameters were selected as $(1,1,1)$ for the non-seasonal structure and $(1,1,1,12)$ for the seasonal part based on both the exploratory analysis and real-world performance of the model. The choice of these parameters was an attempt to achieve a good compromise between model complexity and predictive accuracy. They determine seasonality at a monthly level while using the Fourier terms to represent the yearly seasonal cycle. After the SARIMA model is trained, its predictions for the training period are utilized to create the residual series. Residuals are the short-term differences that SARIMA couldn't account for. These residuals form the target dataset for the LSTM model.

The LSTM model was developed by us in TensorFlow. It features a stacked architecture with two LSTM layers, one with sixty four hidden units and the other with thirty two. A single value for the residual forecast is generated by a dense layer. The training of the model is done with the Adam optimizer that has a learning rate of 0.001 and the mean absolute error is used as the loss function since it is resistant to outliers. Training is carried out on sliding-window sequences of fourteen days. Besides temperature, dew point, visibility, wind speed, differenced pressure, and both Fourier features, each input window also contains data for the multivariate structure. In this way, the LSTM will learn not only the interrelation of different weather variables but also the fact that weather changes rapidly if temperatures change.

The forecasting part brings about a big challenge as the model should carry out predictions for 293 days consecutively without being given any real future values. This means a recursive forecasting strategy whereby each predicted value is to be used as input for the next prediction. The LSTM part of the model could, without intervention, give bigger correction values, which would in turn lead to temperature predictions going beyond the ranges of reality. The system addresses this issue by implementing a stable recursive forecasting framework. During the first five days of forecasting, LSTM residual predictions are employed directly to correct the SARIMA baseline. From the sixth day, however, every new residual prediction is assigned a decay factor of 0.92. The decay gradually reduces the LSTM's effect as the horizon gets longer. This averts deviations from becoming too large and ensures that it is finally the SARIMA baseline that controls the long-term trend.

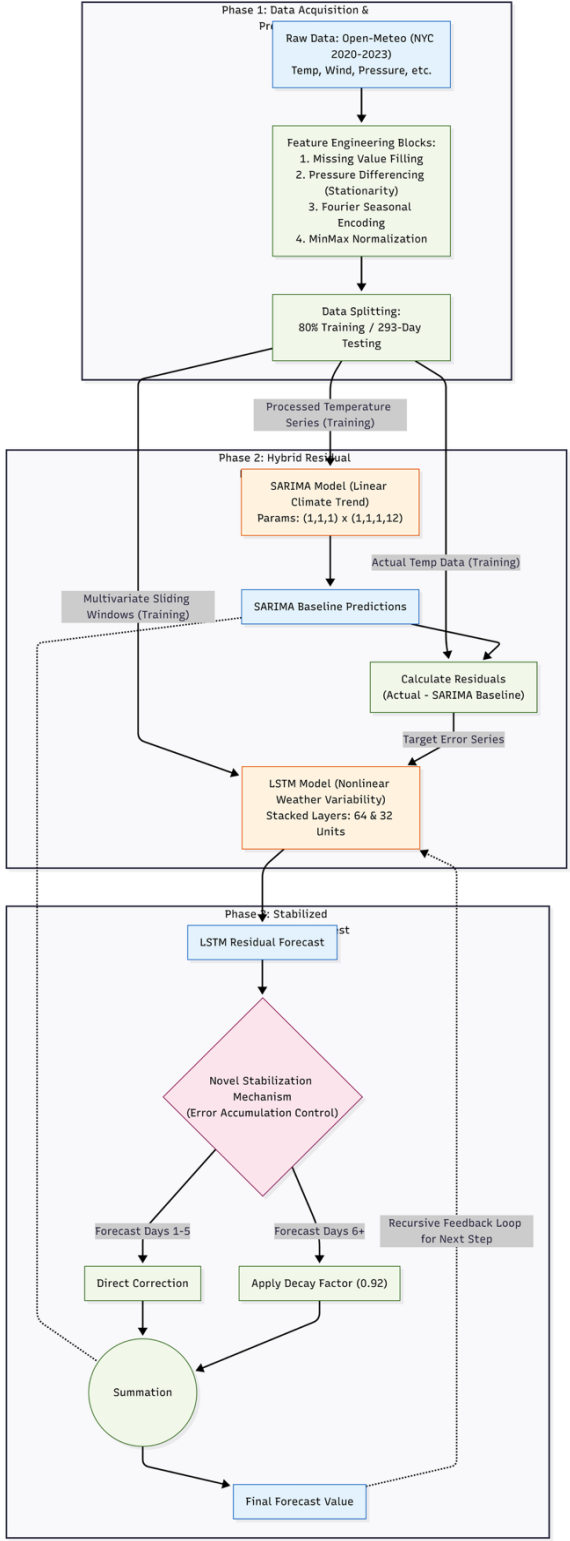

*Figure 1 Architecture of the Model*

This stabilization method is the latest enhancement of the hybrid residual-learning model. It is a compromise between the statistical and neural models that have contradictory strengths. SARIMA is the one that provides you with structured long-term coherence while LSTM is the one that gives you the possibility of making flexible short-term changes. The stabilization step ensures that the LSTM's flexibility doesn't wreak havoc with the forecast in the absence of real future observations. The final model can make multi-month forecasts that are not only accurate but also structurally consistent. This is a very significant improvement compared to standalone LSTM models that quickly diverge and standalone SARIMA models that are unable to capture nonlinear atmospheric changes.

*D. Novel Solution*

The primary novelty of this research lies in the integration of three methodological innovations within a unified forecasting framework: Fourier seasonal encoding, residual-based LSTM training, and a stabilized recursive forecasting mechanism. While hybrid SARIMA–LSTM models have been proposed in the literature, the present study advances this paradigm by addressing the instability typically associated with long-term recursive predictions.

The use of Fourier seasonal encoding enables the model to represent the annual temperature cycle explicitly without relying on SARIMA's seasonal differencing at lag 365, which would be computationally infeasible. This encoding captures long-term periodicity while allowing SARIMA and LSTM to focus on more localized patterns within the time series. The residual-learning approach forces the LSTM to learn only the nonlinear variations that SARIMA cannot model, preventing redundancy and improving efficiency. Most significantly, the introduction of a decay factor into the recursive forecasting loop is a novel stabilization technique that mitigates error accumulation over long prediction horizons. This mechanism ensures that the LSTM contributes meaningfully to the short term without overpowering the SARIMA baseline in the long term.

Together, these innovations produce a model capable of generating accurate, stable, and interpretable forecasts across nearly ten months of future data—an achievement that neither SARIMA nor LSTM could accomplish independently. The hybrid design, informed by principles of statistical modeling, deep learning, and time-series stability, represents a novel and effective solution to the challenges of long-term temperature forecasting.

## IV. RESULTS

The last 293 days from the data collection were utilized to verify the forecasting models. The time span was not employed to train the models so that the output would be fair and unbiased. Each model was required to operate in a fully recursive forecasting environment for this test horizon, which implied that predictions for the earlier days were used as inputs for the later forecast steps. Such a scenario is a better way to forecast the future than one-step-ahead evaluation and it is a very severe test of long-term stability and error accumulation. The same experimental conditions were used to test the Hybrid SARIMA–LSTM model, the independent SARIMA baseline, and the independent LSTM model. We examined their outputs both numerically and qualitatively. For numerical analysis, we resorted to the Mean Absolute

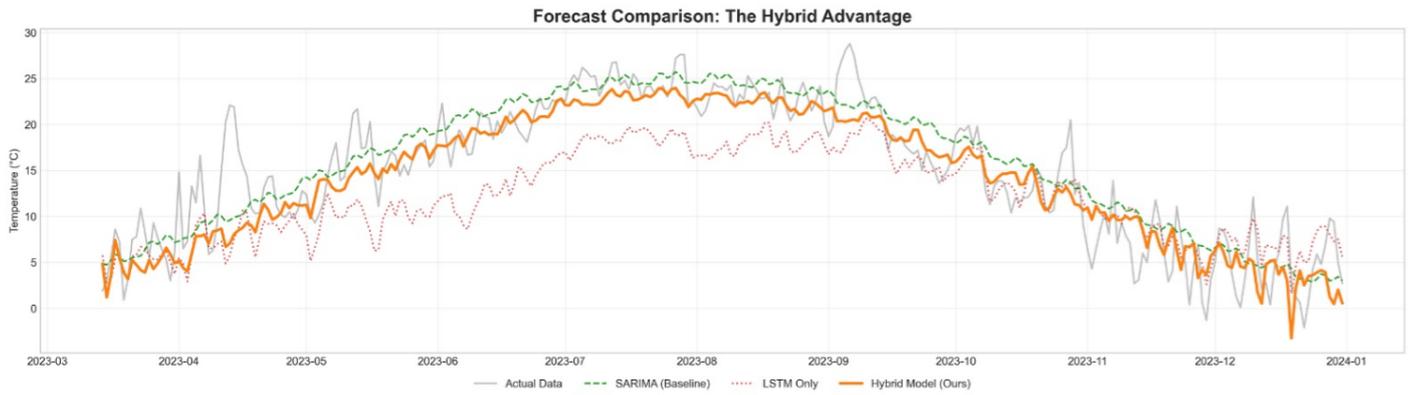

*Figure 2 Forecast Comparison*

Error (MAE) and the Root Mean Square Error (RMSE). As for the qualitative analysis, we implemented accurate forecasts, error growth, and error distribution graphs.

The Hybrid SARIMA–LSTM model demonstrated the most significant overall performance when both numerical and visual tests were utilized. In terms of figures, the hybrid model outperformed the SARIMA baseline with an MAE of 2.41°C and an RMSE of 3.33°C. The SARIMA baseline had an MAE of 2.65°C and an RMSE of 3.34°C. At first, the numerical improvement may seem minor, but there are quite a few reasons why it is significant. As the weather changes quickly and seasonally, temperature forecasting is always a bit of a mistake. A reduction of just 0.20°C in mean absolute error over almost 300 days of forecasts is indicative of a significant improvement in the identification of nonlinear patterns. In addition, a lower RMSE signals better performance in cases of large errors, which are particularly difficult to predict over a long period. Most importantly, the hybrid model is not only making things more accurate, but it also maintains structural realism and avoids catastrophic divergence, which is the main point of this approach.

The SARIMA model alone provided a stable but somewhat limited view of how the temperature would change in the future. The forecast curve of its model followed the expected seasonal drop from summer highs to winter lows, which is what climatologists anticipate for New York City. However, SARIMA always predicted warm anomalies incorrectly and cold anomalies too high. This occurred because SARIMA models only consider linear seasonal patterns and cannot quickly react to short-term weather events. Therefore, the temperature values predicted by its model were usually smoothed over for the most part of the real temperature series, thus missing the important changes. In the long horizon forecast plot, this appears as a baseline trajectory that follows the general shape of the observed temperatures but doesn't fit the local highs and lows. The residual distribution reflects this behavior with SARIMA errors demonstrating a wider spread that is slightly off zero. This indicates both systematic bias and lower precision compared to the hybrid model.

The LSTM model on its own was the worst of the three methods. Its MAE of 8.01°C and RMSE of 9.32°C clearly demonstrate that it was unable to make consistent predictions over the long term. The examination of the predicted route reveals that the first several weeks' predictions are still within reasonable temperature ranges. However, from around day 80 to 100 of the forecast sequence, the LSTM's predictions become quite different from the actual values. This divergence accelerates very quickly, and eventually, the predicted temperatures deviate from the actual temperature curve a lot either above or below. This instability is an issue that has been identified frequently with neural networks that are set up to make recursive predictions. Unless there are structural constraints or stability-preserving mechanisms, the network will use its own predictions as inputs, which can cause small errors to grow over time. Although the LSTM can learn nonlinear relationships during training, it cannot use them for prediction when it does not get the ground truth as feedback. The cumulative error curve is in agreement with this behavior because the LSTM cumulative error increases at an accelerating rate instead of a straight line, thus indicating how errors accumulate.

The complete forecast comparison figure is a very effective tool in revealing these differences between methods. For example, the predicted curve of the hybrid model almost goes hand-in-hand with the real temperature curve for the entire 293 days period. It certainly captures the major seasonal fluctuation, but it also reacts to the short-term changes in a manner that SARIMA cannot. For example, if real temperatures suddenly increase in early autumn or there are short warm spells in late fall, the hybrid model does corrections by the right amount because of LSTM residual predictions. These corrections allow the hybrid model to reflect temporary fluctuations in the weather while still maintaining the overall structure. Conversely, the SARIMA baseline just illustrates a gradual seasonal decline, hence, it misses both the timing and the magnitude of the changes. The LSTM curve, on the other hand, noticeably diverges which, in turn, results in improbable temperature values as residual propagation is not regulated.

The cumulative error graph reveals how the hybrid model is superior to the others even more vividly. The hybrid model is accumulating errors at the slowest pace throughout the entire forecasting period, thus indicating that its predictive behavior over the long term is stable. The SARIMA model, on the other hand, is accumulating errors at a rate that is slightly higher but stable, thus indicating that it is dependable but not very adaptable. However, the LSTM model is characterized by an exponential increase in cumulative error starting from about a third of the way through the horizon. Such a growth pattern is an indication that LSTM is not only inaccurate but also intrinsically unstable during long-range recursive operation. Therefore, although LSTM models may be better at short-term nonlinear prediction, they are not suitable for long-term autonomous forecasting unless they are modified in some way.

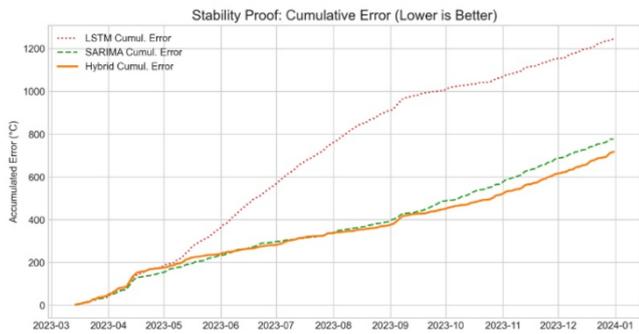

Figure 3 Cumulative Error

The error distribution plot is another piece of evidence that the hybrid approach is the right one. The hybrid residuals are closer to zero and have less spread than the SARIMA residuals. Such a distribution implies that the hybrid model not only has fewer errors, but the errors it makes are more consistent and do not differ significantly in one direction. SARIMA residuals are wider and a little biased as they do not handle nonlinear deviations well. The LSTM residuals are so widely spread that they point to instability, which is in line with the numerical evaluation, to a greater extent.

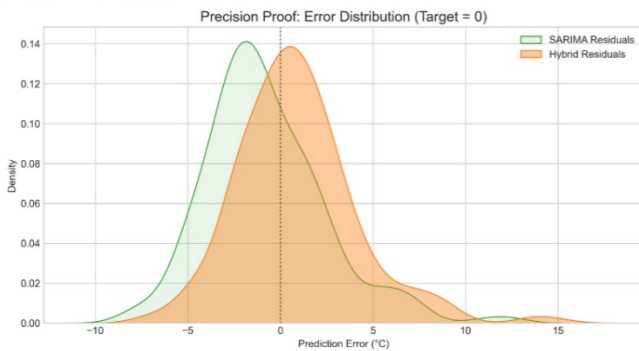

Figure 4 Error Distribution

As a result of these analyses, we can see the enormous potential of the hybrid SARIMA–LSTM model. SARIMA can keep the long-term seasonal trend, while LSTM can very quickly follow short-term and nonlinear changes. The new residual decay factor, the device for stabilization, is very significant in stopping the aggravation of errors. If it weren't for this stabilization, the LSTM part would lower the stability, just like the LSTM model alone. The hybrid model's achievement is an indication that residual learning, when supported by structural constraints and periodic encoding, is a powerful and trustworthy method for long-term temperature prediction.

## V. DISCUSSION

This study's results provide a wealth of understanding of a wide range of pros and cons of statistical, neural, and hybrid forecasting models from the perspective of long-term temperature prediction. The hybrid SARIMA–LSTM model's excellent performance is not only due to its numerical accuracy, but also it reflects, on a deeper level, the compatibility of the two models working together. Temperature time series reveal to us two quite different natures of the same phenomena: on one hand, a slowly changing seasonal cycle resulting from the planet and climate processes, and on the other, fast, random changes caused by short-term atmospheric events. A model that is going to forecast temperature for a long period has to be capable of demonstrating both these behaviors simultaneously.

The hybrid architecture achieves this by assigning the foreseeable parts to SARIMA and the unpredictable parts to LSTM. It allows each sub-model to work on the patterns it is best able to learn. Perhaps the most significant thing to be aware of is that SARIMA alone produces a prediction curve that is stable but overly smoothed. This is what SARIMA does: it makes its prediction by looking at past seasonal and autoregressive patterns, which generally have a smoothing effect on short-lived changes. But, at the same time, these changes in temperature are frequently the ones that tell us the most about short-term weather systems, if we are to forecast the weather.

The hybrid model gets rid of the problems of both SARIMA and LSTM by using their advantages in a residual learning framework. The neural network through the LSTM being trained on the residuals after fitting SARIMA is made to focus only on short-term changes. This clarifies the division of labor: SARIMA takes care of the long-term structure of the climate, while LSTM captures the non-linear changes in the weather. The LSTM is no longer required to learn seasonality, trends, or stationarity. These are all things that neural networks struggle with over long periods. Instead, it learns how the weather changes impact the difference from the expected baseline. This arrangement is not only more interpretable but also makes training more efficient since the network does not have to expend capacity learning patterns that SARIMA has already learned.

The main reason for the hybrid model's effectiveness is the residual decay mechanism introduced during recursive forecasting. If this stabilization were not there, as the forecast horizon got longer, the LSTM would have had more of an effect on the predictions until it exhibited the same divergence behavior as the standalone LSTM model. By gradually reducing the size of the residual corrections after the first few days, the model ensures that the LSTM's nonlinear adjustments remain significant in the short term but lose their power as prediction uncertainty increases. This resembles the situation in operational climate models, where short-term forecasts rely heavily on dynamic simulations and longer-term forecasts depend more on climatological averages. The decay factor ensures that the model stays on SARIMA's stable seasonal path so that the predictions are still there even when it is close to the end of the 293-day forecast window.

Also, Fourier seasonal encoding is very crucial for the model's performance. By using sinusoidal functions to depict the annual cycle, the model obtains a clear and uninterrupted view of seasonal changes. In this way, the model is not allowed to depend only on lagged values for seasonality. For SARIMA, Fourier terms simplify and stabilize seasonal differencing. As for LSTM, Fourier terms deepen the concept of temporal position within the year, thus enabling the network to differentiate between seasonality-caused anomalies in different seasons even if the temperature values are the same. For example, a temperature of 10°C in early spring and one in late fall do not signify the same thing about the atmosphere. Fourier terms allow the model to recognize this difference.

The hybrid SARIMA–LSTM approach is the most effective with meteorological data as it regards temperature both as a seasonal cycle inherently subject to variation and as a volatile atmospheric variable. This dual viewpoint clarifies the performance of the proposed method and, furthermore, reveals the vast potential of hybrid models for climate and environmental forecasting.

## VI. LIMITATIONS & FUTURE WORK

This study's results provide a wealth of understanding of a wide range of pros and cons of statistical, neural, and hybrid forecasting models from the perspective of long-term temperature prediction. The hybrid SARIMA–LSTM model's excellent performance is not only due to its numerical accuracy, but also it reflects, on a deeper level, the compatibility of the two models working together.Temperature time series reveal to us two quite different natures of the same phenomena: on one hand, a slowly changing seasonal cycle resulting from the planet and climate processes, and on the other, fast, random changes caused by short-term atmospheric events. A model that is going to forecast temperature for a long period has to be capable of demonstrating both these behaviors simultaneously.

The hybrid architecture achieves this by assigning the foreseeable parts to SARIMA and the unpredictable parts to LSTM. It allows each sub-model to work on the patterns it is best able to learn. Perhaps the most significant thing to be aware of is that SARIMA alone produces a prediction curve that is stable but overly smoothed. This is what SARIMA does: it makes its prediction by looking at past seasonal and autoregressive patterns, which generally have a smoothing effect on short-lived changes. But, at the same time, these changes in temperature are frequently the ones that tell us the most about short-term weather systems, if we are to forecast the weather.

The hybrid model gets rid of the problems of both SARIMA and LSTM by using their advantages in a residual learning framework. The neural network through the LSTM being trained on the residuals after fitting SARIMA is made to focus only on short-term changes.This clarifies the division of labor: SARIMA takes care of the long-term structure of the climate, while LSTM captures the non-linear changes in the weather. The LSTM is no longer required to learn seasonality, trends, or stationarity. These are all things that neural networks struggle with over long periods. Instead, it learns how the weather changes impact the difference from the expected baseline. This arrangement is not only more interpretable but also makes training more efficient since the network does not have to expend capacity learning patterns that SARIMA has already learned.

The main reason for the hybrid model's effectiveness is the residual decay mechanism introduced during recursive forecasting. If this stabilization were not there, as the forecast horizon got longer, the LSTM would have had more of an effect on the predictions until it exhibited the same divergence behavior as the standalone LSTM model. By gradually reducing the size of the residual corrections after the first few days, the model ensures that the LSTM's nonlinear adjustments remain significant in the short term but lose their power as prediction uncertainty increases. This resembles the situation in operational climate models, where short-term forecasts rely heavily on dynamic simulations and longer-term forecasts depend more on climatological averages. The decay factor ensures that the model stays on SARIMA's stable seasonal path so that the predictions are still there even when it is close to the end of the 293-day forecast window.

Also, Fourier seasonal encoding is very crucial for the model's performance. By using sinusoidal functions to depict the annual cycle, the model obtains a clear and uninterrupted view of seasonal changes. In this way, the model is not allowed to depend only on lagged values for seasonality. For SARIMA, Fourier terms simplify and stabilize seasonal differencing. As for LSTM, Fourier terms deepen the concept of temporal position within the year, thus enabling the network to differentiate between seasonality-caused anomalies in different seasons even if the temperature values are the same. For example, a temperature of 10°C in early spring and one in late fall do not signify the same thing about the atmosphere. Fourier terms allow the model to recognize this difference.

In a broader sense, these results suggest that hybrid methods are inherently the best choice for weather forecasting scenarios where the data reflect both are smooth seasonal cycles and have some short-term changes that are difficult to predict. The hybrid model's structure is, in fact, the most advantageous for long-term recursive forecasting as it brings together stability, adaptability, and interpretability in a single system. The SARIMA baseline is what grounds the model in climate data, the LSTM provides the necessary flexibility to account for atmospheric changes, and the decay mechanism is there to make sure that the predictions are still believable over a long period of time.

The hybrid SARIMA–LSTM approach is the most effective with meteorological data as it regards temperature both as a seasonal cycle inherently subject to variation and as a volatile atmospheric variable. This dual viewpoint clarifies the performance of the proposed method and, furthermore, reveals the vast potential of hybrid models for climate and environmental forecasting.

## VII. CONCLUSION

This research designed a hybrid SARIMA-LSTM forecasting framework that can generate precise and steady long-term predictions of daily temperature in New York City. The temperature series was broken down into a linear seasonal component and a nonlinear residual component, so the model merges the interpretability and structural reliability of SARIMA with the adaptability of LSTM in capturing short-term atmospheric variability.

By implementing Fourier seasonal encoding and a new stabilization method for recursive forecasting, the system was able to keep its predictions consistent over a 293-day horizon and thus, it was more accurate and stable than both the individual models. The findings indicate that hybrid residual-learning strategies constitute a promising and efficient solution to the problem of environmental time-series forecasting, especially when it is necessary to represent both long-range climatological behavior and dynamic short-term anomalies.

In sum, the envisaged model serves as a stepping stone towards more sophisticated hybrid climate forecasting systems and it underscores the necessity of integrating statistical structure with deep learning potentialities in prediction tasks with a long horizon.

REFERENCES

[1] [W. Cao, H. Li, and Y. Sun, "Using SARIMA and LSTM Models to Forecast the Temperature of Internal Walls of Building Facades: A Case Study of Residential Buildings in Xi'an," in Proc. 37th Chinese Control and Decision Conf. (CCDC), Xiamen, China, 2025, pp. 1635–1640. doi: 10.1109/CCDC65474.2025.11090431.

[2] S. Kumari and P. Muthulakshmi, "SARIMA Model: An Efficient Machine Learning Tool for Weather Forecasting," Procedia Computer Science, vol. 235, pp. 656–670, 2024.

[3] L. He, Y. Zhang, Q. Li, J. Zhao, H. Wang, and T. Zhang, "Deep Learning-based Feature Importance for Rainfall Nowcast based on GNSS PWV and CAPE," IEEE J. Sel. Top. Appl. Earth Obs. Remote Sens., vol. 18 no. 1, pp. 26688–26698, 2025. doi: 10.1109/JSTARS.2025.3621857.

[4] M. L. Hossain, S. M. N. Shams, and S. M. Ullah, "Time-Series and Deep Learning Techniques for Renewable Energy Forecasting in Dhaka: A Comparison among three models (ARIMA, SARIMA and LSTM)," Discover Sustainability, vol. 6, art. no. 775, 2025.

[5] E. Krascsenits and A. Kiss, "Comparative Analysis of ARIMA and LSTM Models for Weather Forecasting Using Time Series Data," *Proc. 23rd Int. Symp. Intell. Syst. Informatics (SISY)*, Subotica, Serbia, pp. 119–124, 2025, doi: 10.1109/SISY67000.2025.11205407.

[6] C. E. Machado, K. R. Da Paixão Ataíde, and V. R. P. Borges, "Long Short-Term Memory Approaches for Weather Forecasting from Local Stations," *Proc. 9th Int. Conf. Frontiers Signal Processing (ICFSP)*, Paris, France, pp. 123–127, 2024, doi: 10.1109/ICFSP62546.2024.10785333.

[7] P. Chen, A. Niu, D. Liu, W. Jiang, and B. Ma, "Time Series Forecasting of Temperatures Using SARIMA: An Example from Nanjing," *IOP Conf. Ser.: Mater. Sci. Eng.*, vol. 394, no. 5, p. 052024, 2018, doi: 10.1088/1757-899X/394/5/052024.

[8] https://open-meteo.com/en/docs/historical-weather-api